\documentclass{article}

\usepackage[final]{nips_2018}

\usepackage[utf8]{inputenc} 
\usepackage[T1]{fontenc}    
\usepackage{hyperref}       
\usepackage{url}            
\usepackage{booktabs}       
\usepackage{amsfonts}       
\usepackage{nicefrac}       
\usepackage{microtype}      
\usepackage{algorithm,algorithmicx,algpseudocode}

\usepackage{trackchanges}

\title{Combining Learned Lyrical Structures and Vocabulary for Improved Lyric Generation}

\author{
  Pablo Samuel Castro\\
  Google Brain \\
  \texttt{psc@google.com} \\
  \And
  Maria Attarian \\
  Google \\
  \texttt{jmattarian@google.com} \\
}

\begin{document}

\maketitle

\begin{abstract}
  The use of language models for generating lyrics and poetry has received an increased interest in the last few years. They pose a unique challenge relative to standard natural language problems, as their ultimate purpose is creative; notions of accuracy and reproducibility are secondary to notions of lyricism, structure, and diversity. In this creative setting, traditional quantitative measures for natural language problems, such as BLEU scores, prove inadequate: a high-scoring model may either fail to produce output respecting the desired structure (e.g. song verses), be a terribly boring creative companion, or both. In this work we propose a mechanism for combining two separately trained language models into a framework that is able to produce output respecting the desired song structure, while providing a richness and diversity of vocabulary that renders it more creatively appealing.
\end{abstract}

\section{Introduction}
\label{sec:intro}
With the increased realism and sophistication of generative models, artists have been increasingly drawn to incorporate these methods into their creative process. The approaches vary, from transferring style from one artist to another \citep{dumoulin16learned} to adapting a pre-existing process to produce abstract art that maximizes the likelihood of a category under a classification model \citep{white18perception}.

Lyrics are a particularly challenging artistic endeavour; high-quality lyrics typically require following a specific lyric structure, the use of a rich vocabulary, a mastery of the language, and the use of poetic techniques such as metaphors and alliteration. Because of this, the use of machine learning models for the generation of lyrics has seen a slower increase. The few cases where machine learning models have been used for lyric generation have required a substantial amount of human intervention. In our submission to the Machine Learning and Creativity Workshop at NIPS 2017 \citep{castro17sparkle} we trained a Recurrent Neural Network (RNN) over a dataset of lyrics. We then manually curated the lyrics produced with renowned Canadian songwriter David Usher to rewrite one of his songs (Sparkle and Shine). Although a successful experiment in human-machine collaboration, the lyrics required more manual intervention than we would have liked. We recently switched to the more sophisticated Transformer Language Models (TLMs) \citep{vaswani17attention} to train over the same dataset. The results are of substantially improved, but although they seem to maintain the general \emph{structure} of lyrics, they still suffer from a lack of variety.

\section{Proposed Framework}
\label{sec:framework}
Our approach combines two different TLMs. The first model ($L_S$) is trained to capture the structure of lyrics, while the second ($L_V$) is trained to provide a richer vocabulary than what is currently available in the lyrics dataset, while still leveraging the context of the existing lyrics. Given an initial input lyric $l_1$, we combine these models to produce the next line as follows ($PoS$ will be described below): $l_2 = RichLyrics(l_1) \equiv L_V(l_1 \cdot L_S(PoS(l_1)))$.

$\mathbf{L_S}$: Our dataset consists of a large set of lyrics spanning multiple genres and decades (see Appendix~\ref{sec:lyrics_dataset}). Our inputs consist of the separate lines of all the song lyrics, while the targets are the same lines shifted by one (e.g. $line_n \rightarrow line_{n+1}$). We pre-processed the lyrics by converting them into their respective Parts-of-Speech (PoS)\footnote{We used $pos\_tag$ from Python's $nltk$ library to extract PoS.}. This was done to ensure that the Lyric model is only capturing lyric structure, but not vocabulary. We will refer to this conversion process as $PoS(l)$; in other words our input-to-target mapping becomes $PoS(line_n)\rightarrow PoS(line_{n+1})$.

$\mathbf{L_V}$: We picked a subset of Project Guttenberg's Top 20 books Kaggle dataset\footnote{https://www.kaggle.com/currie32/project-gutenbergs-top-20-books} (the list of books used is provided in Appendix~\ref{sec:books_dataset}). We split each sentence $s$ into two parts: $s_1$ and $s_2$; the splitting point was chosen at about halfway through the sentence, without splitting any words (see Appendix~\ref{sec:word_splitting} for more details). Denoting $\cdot$ as string concatenation, a sentence $s$ is converted to an input-to-target mapping as: $s_1 \cdot PoS(s_2)\rightarrow s_2$. The intuition behind this approach is that $s_1$ provides the \emph{context}, while $PoS(s_2)$ provides the \emph{structure} to be ``materialized''.

\section{Empirical Evaluation}
In order to evaluate our approach we generated 100 lyric verses using the following procedure. We randomly picked 100 lines from our lyrics dataset as starter lines $l_1$. Then, for each model we incrementally built a verse of $5$ lines by setting $l_{i+1}\gets RichLyrics(l_i)$. We are using beam search with max size 3, so each $l_i$ results in 3 different $l_{i+1}$s. We consider the verse produced by each of these possibilities. This means that for each $l_1$ we produce up to $3^5 = 243$ different verses (depending on the input, the beam search may sometimes produce less than 3 variants). We compared our $RichLyrics$ approach against two baselines: $PureLyrics$ is a TLM trained only on the lyrics dataset; $PureBooks$ is a TLM trained only on the books dataset.

\subsection{Quantatitative Evaluation}
From the verses generated by each model we computed the number of words and average word length per line, the number line repeats in the verse ($l_i\equiv l_{i+1}$), and the fraction of words repeated from one line to the next. The results are presented in Table~\ref{tbl:stats} and demonstrate that $RichLyrics$ makes use of a much larger vocabulary and with fewer repeats. Given that the verses are 5 lines long, $PureLyrics$ is repeating lines about half the time! Qualitative examples of the generations are presented in Appendix~\ref{sec:qualitative} and further confirm these quantitative results.
\begin{table}[h]
  \begin{tabular}{|l|l|l|l|l|}
    \hline
    & \textbf{Words per line} & \textbf{Word length} & \textbf{Number of line} & \textbf{Fraction of} \\
    &                         & \textbf{per line}    & \textbf{repeats}        & \textbf{repeated words} \\
    \hline
    $PureLyrics$ & $12.853\pm 1.106$ & $3.351\pm 0.060$ & $2.233\pm 0.125$ & $0.671\pm 0.026$ \\
    \hline
    $PureBooks$ & $6.633\pm 0.024$ & $3.758\pm 0.010$ & $0.002\pm 0.001$ & $0.244\pm 0.002$ \\
    \hline
    $RichLyrics$ & $6.928\pm 0.046$ & $3.542\pm 0.016$ & $0.441\pm 0.039$ & $0.480\pm 0.009$ \\
    \hline
  \end{tabular}
  \caption{Statistics for the different models.}
  \label{tbl:stats}
\end{table}

\section{Discussion and Future Work}
Although we are able to substantially improve the quality of the generated lyrics, there is still much work ahead of us. We would like to train over a larger set of books, and ones that are more current to have more modern vocabulary. An important aspect of lyric structure that we are investigating is having the generation adapt to rhyming structure and phonetic cadence, as this is something songwriters use often to fit a musical melody. As with most language models out there, semantic consistency still proves challenging, and is something we are actively investigating.

\newpage

\bibliographystyle{apalike}
\bibliography{rich_lyrics}

\appendix
\section{Lyrics dataset}
\label{sec:lyrics_dataset}
These are the genres used for the lyric structure model detailed in Section~\ref{sec:framework}. We exclude Children's Music and Hip-Hop, the latter to reduce the amount of profanity in the generations.
\begin{itemize}
  \item Alternative / indie
  \item Country
  \item Folk
  \item Jazz
  \item Metal
  \item Pop
  \item R-and-B / Soul
  \item Rock
  \item Soundtracks
\end{itemize}

In total this resulted in over 1 million input/target pairs, and about 91,000 for the test/validation sets.

\section{Kaggle books dataset}
\label{sec:books_dataset}
The books used for training the Lyric Vocabulary model detailed in Section~\ref{sec:framework} were:
\begin{itemize}
    \item A Tale of Two Cities by Charles Dickens
    \item Adventures of Huckleberry Finn by Mark Twain
    \item Alices Adventures in Wonderland by Lewis Carroll
    \item Dracula by Bram Stoker
    \item Emma by Jane Austen
    \item Frankenstein by Mary Shelley
    \item Great Expectations by Charles Dickens
    \item Grimms Fairy Tales by The Brothers Grimm
    \item Metamorphosis by Franz Kafka
    \item Pride and Prejudice by Jane Austen
    \item The Adventures of Sherlock Holmes by Arthur Conan Doyle
    \item The Adventures of Tom Sawyer by Mark Twain
    \item The Count of Monte Cristo by Alexandre Dumas
    \item The Picture of Dorian Gray by Oscar Wilde
    \item The Prince by Nicolo Machiavelli
    \item The Yellow Wallpaper by Charlotte Perkins Gilman
\end{itemize}

In total this resulted in around 155,000 input/target pairs used for training, and around 13,000 for the test/validation sets.

\section{Word splitting mechanism}
\label{sec:word_splitting}
The split happens assuming that periods denote the end of a sentence (aside from abbreviations typically containing a period e.g. Mr. or St.) while also taking into account that quotes in books sometimes mark the end of a sentence, e.g. in the case of quotes being followed by an uppercase word. In order to avoid very long sentences skewing the structure of the proposed framework, we are splitting into subsentences of 15 words each, if the total word number in a sentence exceeds a threshold. Subsequently, we split each sentence into approximately half, in respect of not cutting words in half. This has been parameterized in case we choose to experiment with different ratios.

As an example, if we have the sentence: ``The quick brown fox jumped over the fence'', our procedure would produce the following input and target phrases:
\begin{itemize}
  \item \textbf{Input:} The quick brown fox VBD IN DT NN
  \item \textbf{Target:} jumped over the fence
\end{itemize}

\newpage

\section{Qualitative Evaluation}
\label{sec:qualitative}
We present some sample lyrics produced by the different models, using the same starter lines (in italics below). As discussed in Section~\ref{sec:framework}, we generate lines incrementally: $l_1\rightarrow l_2\rightarrow\cdots l_n$. In Table~\ref{tbl:lyrics_v_rich} we compare $PureLyrics$ with $RichLyrics$, where the increased variety in outputs produced by $RichLyrics$ is evident.

\begin{table}[h]
  \caption{Comparison between $PureLyrics$ and $RichLyrics$ models.}
  \label{tbl:lyrics_v_rich}
  \begin{tabular}{|c|c|}
    \hline
    \textbf{PureLyrics} & \textbf{RichLyrics} \\
    \hline
    \emph{come on , uh} & \emph{come on , uh} \\
    i'm not gonna write you a love song & you remember the voice of the widow \\
    'cause you tell me it's & i love the girl of the age \\
    i'm not the one you wanted & i have a regard for the whole \\
    i'm not the one you wanted & i have no doubt of the kind \\
    i'm not the one you wanted & i am sitting in the corner of the mantelpiece \\
    \hline
    \emph{i 'm your big and brave and handsome romeo} & \emph{i 'm your big and brave and handsome romeo} \\
    i'm the man of the woods,  i'm the man of the woods & you know my secret secret \\
    i'm the man of the woods,  i'm the woods & you have my second estate \\
    i'm the man of the woods,  i'm the woods & you suit your high origin \\
    i'm the man of the woods,  i'm the woods & you have my cursed youth \\
    i'm the man of the woods,  i'm the woods & you have my life \\
    \hline
    \emph{you told me you loved nobody else , you never would} & \emph{you told me you loved nobody else , you never would} \\
    you told me you loved nobody else but you & you told me you wanted everything else, you never would \\
    you told me you loved nobody but you & he put his hand on the pillow of the marquis \\
    you told me you loved nobody but you & he put his cap on the ground like a stone \\
    you told me you loved nobody but you & he put his hand on the latch of a door \\
    you told me you loved me but you loved me & he put his key in the lock as a key \\
    \hline
  \end{tabular}
\end{table}

In Table~\ref{tbl:books_v_rich} we compare $PureBooks$ with $RichLyrics$, which highlights how our proposal produces output that is more reminiscent of real lyrics, both in terms of phrase structure and length.

\begin{table}[h]
  \caption{Comparison between $PureBooks$ and $RichLyrics$ models.}
  \label{tbl:books_v_rich}
  \begin{tabular}{|c|c|}
    \hline
    \textbf{PureBooks} & \textbf{RichLyrics} \\
    \hline
    \emph{come on , uh} & \emph{come on , uh} \\
    she was a new man & you remember the voice of the widow \\
    but it was not & i love the girl of the age \\
    a thing to be done & i have a regard for the whole \\
    it was & i have no doubt of the kind \\
    a confession & i am sitting in the corner of the mantelpiece \\
    \hline
    \emph{i 'm your big and brave and handsome romeo} & \emph{i 'm your big and brave and handsome romeo} \\
    you, and i'll tell you all about it & you know my secret secret \\
    i don't & you have my second estate \\
    understand you, & you suit your high origin \\
    said the young man, and we & you have my cursed youth \\
    shall be happy to-day & you have my life \\
    \hline
    \emph{you told me you loved nobody else , you never would} & \emph{you told me you loved nobody else , you never would} \\
    have felt that you were coming to know of yourself & you told me you wanted everything else, you never would \\
    i have no & he put his hand on the pillow of the marquis \\
    doubt of that, & he put his cap on the ground like a stone \\
    said the young man, that you have been a & he put his hand on the latch of a door \\
    great fancy for a few minutes, and then another? & he put his key in the lock as a key \\
    \hline
  \end{tabular}
\end{table}

\end{document}